\documentclass[letterpaper]{article} 
\usepackage[submission]{aaai23}  
\usepackage{times}  
\usepackage{helvet}  
\usepackage{courier}  
\usepackage[hyphens]{url}  
\usepackage{graphicx} 
\usepackage{natbib}  
\usepackage{caption} 
\usepackage{algorithm}
\usepackage[noend]{algorithmic}

\usepackage{newfloat}
\usepackage{listings}
\DeclareCaptionStyle{ruled}{labelfont=normalfont,labelsep=colon,strut=off} 
\floatstyle{ruled}
\newfloat{listing}{tb}{lst}{}
\floatname{listing}{Listing}
\usepackage{booktabs}       
\usepackage{amsfonts}       

\usepackage{multirow}
\usepackage{subcaption}

\usepackage{amsmath,amsfonts,bm}

\def\eqref#1{equation~\ref{#1}}

\def\1{\bm{1}}

\DeclareMathAlphabet{\mathsfit}{\encodingdefault}{\sfdefault}{m}{sl}
\SetMathAlphabet{\mathsfit}{bold}{\encodingdefault}{\sfdefault}{bx}{n}

\usepackage{thm-restate}

\newcommand{\method}{RegMix}

\title{\method{}: Data Mixing Augmentation for Regression}
\author {
    Seong-Hyeon Hwang,
    Steven Euijong Whang
}
\affiliations {
    KAIST\\
    \{sh.hwang, swhang\}@kaist.ac.kr
}

\usepackage{bibentry}

\begin{document}

\maketitle

\begin{abstract}
Data augmentation is becoming essential for improving regression performance in critical applications including manufacturing, climate prediction, and finance. Existing techniques for data augmentation largely focus on classification tasks and do not readily apply to regression tasks. In particular, the recent Mixup techniques for classification have succeeded in improving the model performance, which is reasonable due to the characteristics of the classification task, but has limitations in regression. We show that mixing examples that have large data distances using linear interpolations may have increasingly-negative effects on model performance. Our key idea is thus to limit the distances between examples that are mixed. We propose \method{}, a data augmentation framework for regression that learns for each example how many nearest neighbors it should be mixed with for the best model performance using a validation set. Our experiments conducted both on synthetic and real datasets show that \method{}  outperforms state-of-the-art data augmentation baselines applicable to regression.
\end{abstract}

\section{Introduction}
\label{sec:introduction}
As machine learning (ML) becomes widely used in critical applications including manufacturing, climate prediction, and finance, data augmentation for regression becomes essential as it provides an opportunity to improve model performance without additional data collection. In comparison to classification tasks such as object detection in images, regression tasks predict real numbers.  

\begin{figure}[t]
\centering
\includegraphics[width=0.88\columnwidth]{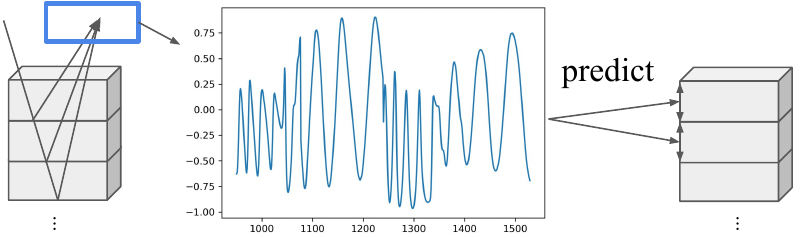}
\caption{Spectrum generation on a 3-d semiconductor. }
\label{fig:spectrumgeneration}
\end{figure}

To emphasize the importance of data augmentation in regression, we provide a case study of semiconductor manufacturing. Here a common quality check is to measure the layer thicknesses of a 3-dimensional semiconductor and see if they are even. However, directly measuring each thickness results in destroying the semiconductor itself, so a recently-common approach is to take an indirect measurement by applying light waves on the semiconductor, measuring the spectrum of wavelengths that bounce back from all the layers, and use ML to predict the layer thicknesses from the spectrum data (see Fig.~\ref{fig:spectrumgeneration} for an illustration). With enough spectrum data and thickness information, ML models can be trained to accurately predict thicknesses from a spectrum. The main challenge is that there is not enough training data, and the only cost-effective solution is to augment small amounts of labeled data that can be produced. According to collaborators in this industry, modeling the physical system using simulators is not practical due to its complexity. Even a small improvement in model performance from data augmentation has significant impact. In general, any regression task (e.g., predicting emissions or stock prices) can benefit from data augmentation.

\begin{figure}[t]
\begin{subfigure}[t]{0.5\columnwidth}
\centering
\includegraphics[scale=0.2]{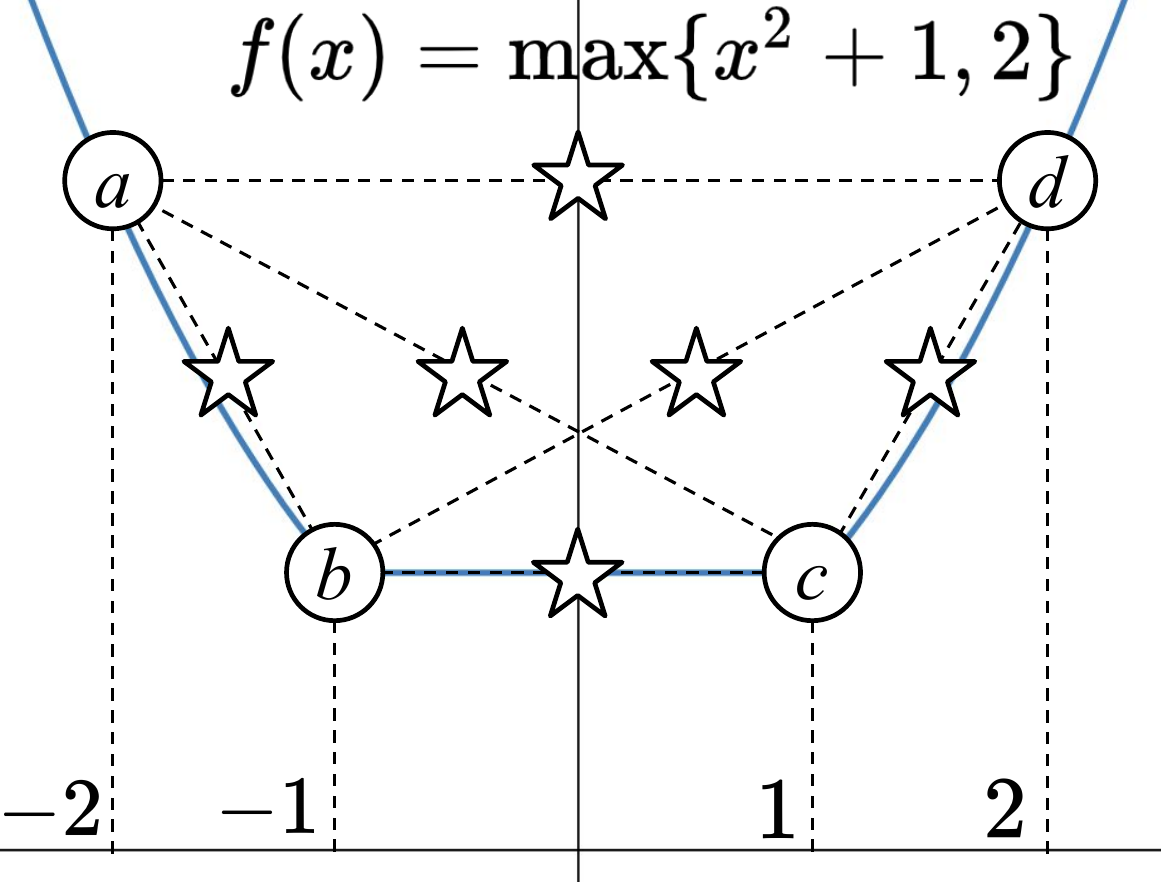}
\caption{}
\label{fig:classificationexample}
\end{subfigure}
\begin{subfigure}[t]{0.45\columnwidth}
\centering
\includegraphics[scale=0.2]{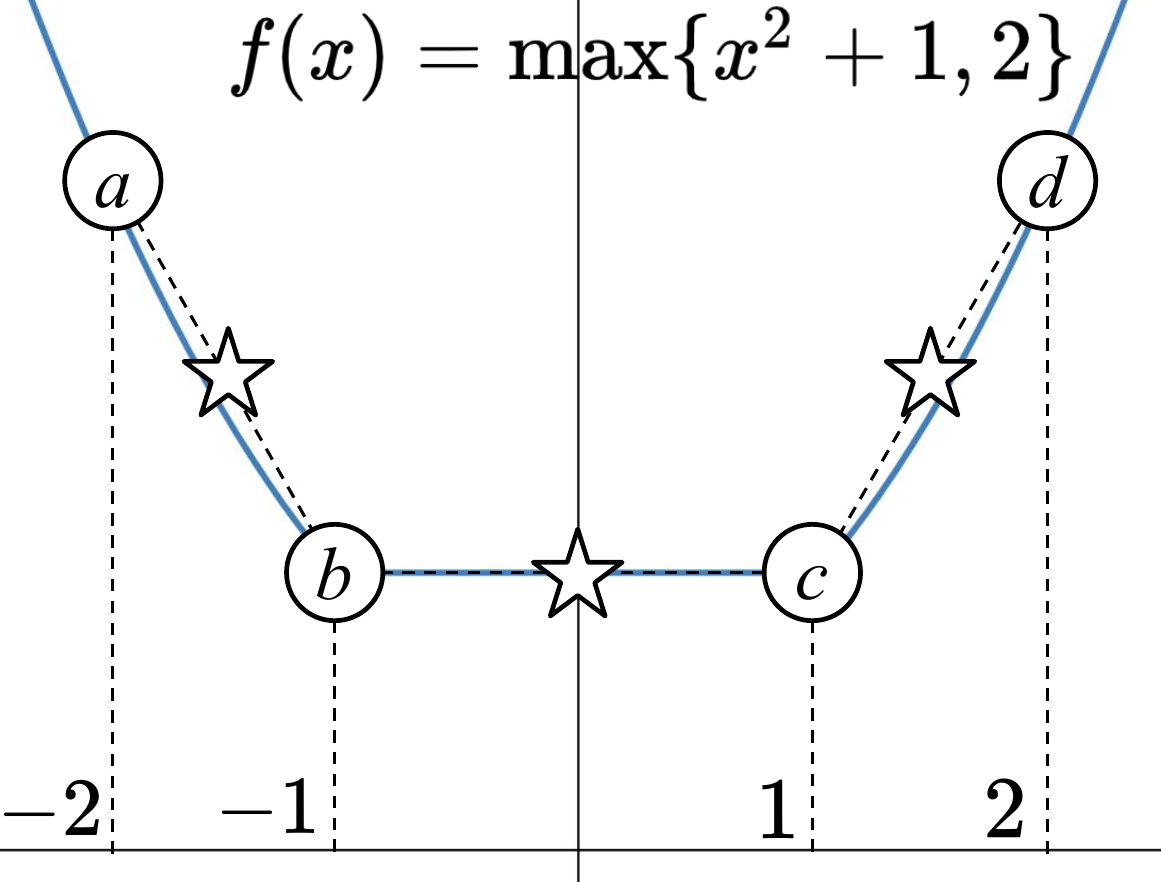}
\caption{}
\label{fig:regressionexample}
\end{subfigure}
  \caption{A comparison between Mixup\,\citep{DBLP:conf/iclr/ZhangCDL18} and \method{} on a small regression dataset where each example (circle) is associated with a single-dimensional feature (x-axis). (a) Mixup takes a linear interpolation of all possible example pairs. Unfortunately, the mixed examples (stars) do not properly reflect the function $f$ (blue plot) where some labels are arbitrarily incorrect. (b) \method{} learns for each example how many nearest neighbors it should be mixed with. As a result, each example is only mixed with its 1- or 2-nearest neighbors to better reflect $f$.}
    \label{fig:comparison}
\end{figure}

Most data augmentation techniques are designed for classification tasks and not necessarily for regression. Data augmentation techniques for classification including image processing (e.g., flipping or rotating) and generative models (e.g., GAN\,\citep{DBLP:conf/nips/GoodfellowPMXWOCB14} and VAE\,\citep{DBLP:journals/corr/KingmaW13}) are not designed for regression tasks (see Sec.~\ref{sec:relatedwork} for more details). More recently, Mixup\,\citep{DBLP:conf/iclr/ZhangCDL18} is a popular data augmentation technique that is primarily used for classification, but can also be used for regression. The idea of Mixup is to mix two examples by linear interpolation and use it to estimate the label of any examples in between. There is theoretical evidence that Mixup regularizes the model being trained\,\citep{DBLP:journals/corr/abs-2006-06049, DBLP:conf/iclr/ZhangDKG021}. Recently, various versions of Mixup\,\citep{verma2019manifold, DBLP:conf/iccv/YunHCOYC19, kim2020puzzle, kim2021comixup} have been proposed to train and calibrate the model better. Although Mixup can be used as is for regression, it is not equally effective because taking linear interpolations may sometimes do more harm than good to the model's performance. The reason is that interpolating examples when the label space is continuous may result in arbitrarily-incorrect labels as illustrated in Fig.~\ref{fig:classificationexample}.


We propose \method{}, a data mixing augmentation framework that is effective for regression tasks by limiting distances between mixing examples. Unlike existing theoretical works on Mixup for regression\,\cite{DBLP:conf/iclr/ZhangDKG021,DBLP:journals/corr/abs-2006-06049,DBLP:conf/icml/WuZVR20}, \method{} assumes that linear interpolations when mixing examples are only effective within certain data distances. The distance limit may vary by example, and we formulate the problem of learning for each example how many nearest neighbors it should be mixed with, which we hereby refer to as a {\em kNN mixing policy}. \method{} employs a framework that searches for these policies that lead to the lowest model losses on a validation set using reinforcement learning (RL)\,\citep{DBLP:conf/iclr/ZophL17}, although other search algorithms like evolution\,\citep{DBLP:conf/icml/HoLCSA19} can be used as well. Our approach is inspired by the AutoAugment framework\,\citep{DBLP:conf/cvpr/CubukZMVL19}, which searches image augmentation policies for classification. A key difference is that AutoAugment searches for the best augmentation operators that are applied to the entire data, while \method{} finds the number of neighbors to mix for each example. Fig.~\ref{fig:regressionexample} shows how limiting the nearest neighbors to mix is better than mixing with all neighbors as in Mixup.

Experiments conducted on real and synthetic datasets show that \method{} outperforms baselines, especially when the mixing must be done selectively. \method{} is designed to improve model performance with more computation than simply mixing examples, so we also suggest using it has a post-hoc method after model training when the extra model performance gain makes a difference.

\section{Limitations of Mixup in Regression}
\label{sec:nonlinear}

In this section, we explain why Mixup in classification has limitations in a regression setting in more detail. The original version of Mixup\,\citep{DBLP:conf/iclr/ZhangCDL18} is to take a linear interpolation between any pair of examples $x_i$ and $x_j$ with the labels $y_i$ and $y_j$ to produce the new example $\lambda x_i + (1 - \lambda) x_j$ with the label $\lambda y_i + (1 - \lambda) y_j$ where $\lambda \sim Beta(\alpha, \alpha)$. According to \citet{DBLP:conf/iclr/ZhangCDL18}, mixing all examples outperforms Empirical Risk Minimization on many classification datasets since the mixed examples have a regularization effect that prevents overfitting. On the other hand, mixing examples may cause manifold intrusion \,\citep{DBLP:conf/aaai/GuoMZ19} where the mixed label is in the wrong label manifold and thus confuses the model.

In regression, however, the degree of manifold intrusion may be worse than classification due to the continuous label space. Since the possible number of label values is infinite, it is easier for a label to be incorrect. And the greater the distance between examples, the more likely the error is significant. For example, mixing the points $a$ and $d$ in Fig.~\ref{fig:classificationexample} results in a label nowhere near the true label. In Sec.~\ref{sec:linearityexperiments}, we observe on real datasets that larger data distances lead to increased label errors and thus worse model performance.

Our key approach is thus to limit the data distance to improve Mixup for regression. If the regression model is continuous and two examples are close enough, the mixed label will be closer to the actual label when using linear interpolation. We provide a simple analysis in the supplementary. Even if the true function is not continuous (e.g., identical examples having different labels), we contend that limiting the data distances for mixing turns out to be an effective solution in practice.

\section{Methodology: \method{}}
\label{sec:mixr}

The goal of \method{} is to identify which examples to mix with which nearest neighbors. Instead of finding the actual distance limits themselves, we solve the identical problem of finding the number of nearest neighbors to mix per example to reduce the search space. A na\"ive solution is to use Bayesian Optimization (BO)\,\citep{10.5555/646296.687872}, but BO is known to be difficult to scale to high dimensions\,\cite{frazier2018tutorial} and is thus not suitable in our setting where the number of parameters is the same as the dataset size. Instead, we use RL to find the policies.

\subsection{Framework}
\label{sec:framework}

Let $\mathcal{D} = \{(x_i, y_i)\}_{i=1}^S \sim P$ be the training set where $x_i \in X$ is a $d$-dimensional input example, and $y_i \in Y$ is an $e$-dimensional label. Let $\mathcal{D}^v = \{(x_i^v, y_i^v)\}_{i=1}^V \sim P^t$ be the validation set, where $P^t$ is the distribution of the test set, which is not necessarily the same as $P$. Let $f_\phi$ be a regression model, and $L$ the loss function that returns a performance score comparing $f_\phi(x_i)$ with the true label $y_i$ using Mean Square Error (MSE) loss. We assume a list $N$ of possible nearest neighbors (NNs) based on data distance that can be mixed with an example. For instance, $N$ could contain the options \{0 NN, 4 NNs, 8 NNs\}. Notice that the ``0 NN'' option means that no mixing occurs because it is not beneficial to model performance. The more fine-grained the kNN options are, the more precisely \method{} can determine the optimal number of NNs to mix per example. On the other hand, there are diminishing returns because the optimization itself may become more difficult (see Sec.~\ref{sec:accuracyresults}). 

\begin{figure}[t]
    \centering
    \includegraphics[scale=0.65]{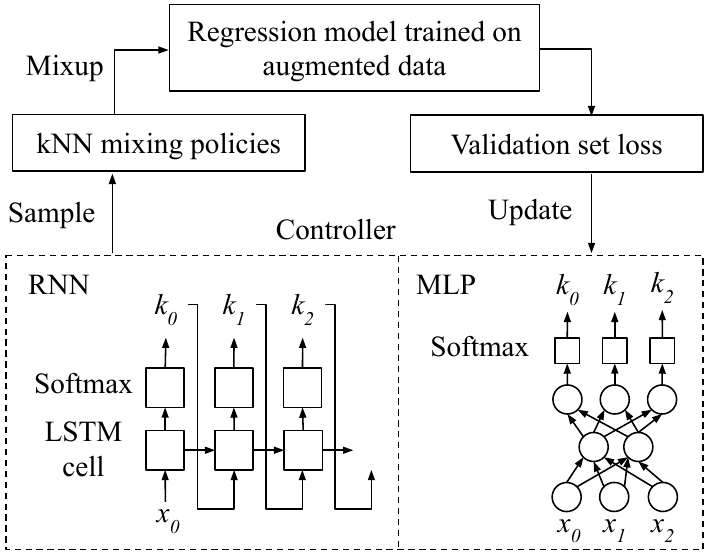}
    \caption{\method{}'s workflow using two possible controllers. Assuming that mixing an example with its neighbors influences subsequent examples, an RNN-based controller can be used to predict the kNN mixing policies for an example sequence. Alternatively, if we do not explicitly model influence, then an MLP-based controller can be used to predict the kNN mixing policies for all examples at once.}
    \label{fig:architecture}
\end{figure}

\method{}'s framework in Fig.~\ref{fig:architecture} is similar to AutoAugment and neural architecture search (NAS)\,\citep{DBLP:conf/iclr/ZophL17} where a controller is trained to generate kNN mixing policies for all the examples in $\mathcal{D}$. The policies are then sampled probabilistically following the distributions of the softmax layers and used to augment $\mathcal{D}$ and train $f_\phi$. The models are then evaluated on $\mathcal{D}^v$, and the losses are used to update the controller. Alg.~\ref{alg:mixr} shows the pseudocode of \method{}. Unlike typical RL settings where an episode consists of a sequence of dependent steps, in our setting an episode consists of independent steps because the controller can reach a goal after taking only one action (i.e., sampling kNN mixing policies). As a result, we can sample policies and compute model losses in parallel (lines 4--8). The collected losses are then used to compute the rewards and update the controller (lines 9--14). 


\begin{algorithm}[H]
\caption{Pseudocode of \method{}.}
\label{alg:mixr}
\textbf{Input}: training set $\mathcal{D}$, validation set $\mathcal{D}^v$, time horizon $T$\\
\textbf{Initialize}: controller $C$, moving average $\delta$ = 0, trajectory $\tau$, loss set $s$\\
\textbf{Output}: kNN mixing policies
\begin{algorithmic}[1] 
    \REPEAT
    \STATE $\tau$ = \{\}, $s$ = \{\};
      \FOR[Can run in parallel]{$i = 1 \ldots T$} 
        \STATE Initialize $\phi$;
        \STATE Sample kNN mixing policies $P$ from $C$\;
        \STATE Train $f_\phi$ with $\phi$ on $\mathcal{D} \cup Mix(\mathcal{D}, P)$\;
        \STATE $Loss = \frac{1}{V}\sum_{i=1}^V L(f_\phi(x_i^v), y_i^v)$\;
        \STATE Add $Loss$ to $s$;
      \ENDFOR
      \FOR{$i = 1 \ldots T$}
        \STATE $Loss$ = $s$[$i$];
        \STATE $r = \frac{1}{Loss}$ - $\delta$;
        \STATE Add $\{P, r\}$ to $\tau$;
        \STATE $\delta$ = 0.95$\delta$ + 0.05$\frac{1}{Loss}$;
      \ENDFOR
        \STATE Update controller with $\tau$;
    \UNTIL{convergence}
\end{algorithmic}
\end{algorithm}


\subsection{Controller Design}
\label{sec:controller}

The controllers in the previous works AutoAugment\,\citep{DBLP:conf/cvpr/CubukZMVL19} and NAS\,\citep{DBLP:conf/iclr/ZophL17} use recursive neural networks (RNNs) because of the sequential nature of the predictions to make. In AutoAugment, a policy consists of a sequence of operations on images that should be performed in a certain order (e.g., equalize and then rotate). In NAS, a model architecture consists of components that are connected in some sequence as well (e.g., elementwise multiplication followed by a ReLU activation function). Using an RNN is thus a natural choice in these cases.

For \method{}'s augmentation, there is no pre-determined sequence among examples, but one example mixing with its neighbors might ``influence'' the kNN mixing policies of other examples. For instance, if there are three examples $a$, $b$, and $c$ that are close together, and $a$ mixes with $b$, then $b$ may not have to mix with $a$ or even $c$ for the best model performance. If the examples can be ordered where examples that influence each other are close by, then an RNN can be used to determine the best kNN values for the sequence of examples. \method{}'s RNN-based controller (Fig.~\ref{fig:architecture}, controller on the left) consists of a sequence of LSTM\,\cite{DBLP:journals/neco/HochreiterS97} cells with softmax layers for the ordered examples similar to AutoAugment and NAS where the result of one cell is used as the input of the next. To find an ordering among examples, one can use dimensionality reduction or any other sorting method.

However, if there is no obvious ordering of examples, then an RNN may not be the ideal architecture. In Sec.~\ref{sec:accuracyresults} and Fig.~\ref{fig:knnhistograms}, we observe that \method{} selectively mixes examples, but there are still quite a few examples that mix with most of the other examples. Mixing such an example may affect many other examples at the same time instead of the next example in some sequence. Here it is better to avoid specifically capturing the direction of influence as in an RNN. 

We thus propose a more light-weight controller based on a multi-layer perceptron (MLP) instead of an RNN (Fig.~\ref{fig:architecture}, controller on the right) where the goal is to predict the best kNN mixing policies of all examples at once. Here the input does not play a role, and we fix it to a vector of 1's. Also, there is one output node with a softmax layer per example. In Sec.~\ref{sec:accuracyresults}, we show that using an MLP controller has similar model performances as when using an RNN controller.


Finding the optimal policy involves taking the gradient of the objective function $J(\theta) = \mathbb{E}_{\pi_\theta}[R]$ where $\pi$ is the policy, $\theta$ is the parameters of the controller, and $R$ is the reward function. We would like to {\em minimize} the regression model loss on a validation set when mixing a batch of examples. However, the validation loss is computed using the regression model, which does not involve $\theta$. Hence we cannot analytically compute the differential of the reward function with respect to $\theta$. In addition, there is no training data on how mixing each example affects the model performance.

We thus use Proximal Policy Optimization (PPO)\,\citep{DBLP:journals/corr/SchulmanWDRK17}, which is one of the state-of-the-art on-policy algorithms and is also used in AutoAugment. We use PPO for convenience, and any actor-critic method like A3C\,\citep{DBLP:conf/icml/MnihBMGLHSK16} can be used as well. In order to minimize the loss, we use the inverse of loss (i.e., $\frac{1}{\text{loss}}$) as the reward function. For stable learning, we employ an exponential moving average of previous rewards as the baseline function to reduce variance. 

\section{Experiments}
\label{sec:experiments}

We provide experimental results for \method{}. We evaluate regression models on separate test sets and repeat all model trainings five times. We use PyTorch\,\citep{paszke2017automatic}, and all experiments are performed using Intel Xeon Silver CPUs and NVidia RTX GPUs. 

\subsection{Experimental Settings}

\paragraph{Measures} 
To measure accuracy, we use RMSE = $\sqrt{\frac{1}{n}\sum^n_{i=1}(y_i-\hat{y}_i)^2}$, which is the absolute difference between predicted labels and true labels where a lower value is better. To measure the distance between examples, we use Euclidean distance.

\paragraph{Datasets}
We use three real and one synthetic datasets. The {\sf NO2} emissions dataset\,\citep{statlib} contains traffic and meteorological information around roads. The features include cars per hour, wind speed, temperature, and others. The label is the NO2 concentration. The {\sf Bike} UCI dataset\,\citep{Dua:2017} contains climate and temporal information to predict the number of bikes in demand where the features include temperature, humidity, rainfall, and season. We take a random subset of the original dataset. The {\sf Product} dataset obtained through a collaboration with a major tech company contains spectrum data that is generated using the procedure described in Sec.~\ref{sec:introduction} and illustrated in Fig.~\ref{fig:spectrumgeneration} on 20-layer 3-d semiconductors. We also use a public synthetic dataset\,\citep{dacon} ({\sf Synthetic}) since the {\sf Product} dataset is proprietary. Here a simulator generates spectrum data using the same procedure as {\sf Product} assuming 4-layer 3-d semiconductors. We provide more details in the supplementary.

Table~\ref{tbl:datasetcomparison} compares the four datasets in more detail. {\sf NO2} and {\sf Product} have the lowest and highest dimensions, respectively, and {\sf Synthetic} has the largest size. Although {\sf Product}'s dimensionality is considered large for applications using spectrum data, there are datasets of even higher dimensions (e.g., large images or spatio-temporal data), and evaluating \method{} on them is an interesting future work. We also note that the data sizes are intentionally small as it is critical to show that \method{} performs well for small datasets. Finally, there are no personal identifiers.


\begin{table}[t]
  \caption{Settings for the four datasets.}
  \centering
  \small
  \begin{tabular}{@{\hspace{4pt}}c@{\hspace{4pt}}c@{\hspace{4pt}}c@{\hspace{4pt}}c@{\hspace{4pt}}c@{\hspace{4pt}}}
    \toprule
    Dataset & Data dim. & Label dim. & Train. set size & Val. set size\\
    \midrule
    {\sf NO2} & 7 & 1 & 200 & 100 \\
    {\sf Bike} & 12 & 1 & 300 & 200 \\
    {\sf Product} & 580 & 20 & 300 & 200 \\
    {\sf Synthetic} & 226 & 4 & 600 & 400\\
    \bottomrule
  \end{tabular}
  \label{tbl:datasetcomparison}
\end{table}

\paragraph{\method{} Settings} 

As a default, we evaluate \method{} using the MLP controller for all results. When using an MLP controller, we use an MLP with 4 hidden layers and set the number of nodes per hidden layer to be [100, 100, 100, 100] for the four datasets with a learning rate of 0.002. When using an RNN controller, we use an embedding layer to adjust the dimension of inputs and LSTM cells with a hidden state size of 100. We use a learning rate of $\alpha = 0.0005$. For the regression model $f_{\phi}$, we use an MLP using hyperparameters shown in Table~\ref{tbl:regmodelsetting}. We also evaluate using other models in the supplementary, and the overall performance results are similar. We use layer normalization\,\citep{DBLP:journals/corr/ba2016layer} for the {\sf Product} and {\sf Synthetic} datasets.

\begin{table}[t]
  \caption{Settings for regression models on the four datasets.}
  \centering
  \small
  \begin{tabular}{@{\hspace{4pt}}c@{\hspace{4pt}}c@{\hspace{4pt}}c@{\hspace{4pt}}c@{\hspace{4pt}}}
    \toprule
    Dataset & MLP structure & Learning rate & Batch size\\
    \midrule
    {\sf NO2} & [512, 256] & 0.0001 & 32 \\
    {\sf Bike} & [512, 256] & 0.0001 & 64 \\
    {\sf Product} & [512, 256, 64] & 0.0001 & 64 \\
    {\sf Synthetic} & [2048, 1024, 512, 256] & 0.001 & 64\\
    \bottomrule
  \end{tabular}
  \label{tbl:regmodelsetting}
\end{table}

When running Alg.~\ref{alg:mixr}, we set the time horizon $T$ to 20 and run lines 4--8 in parallel using multi-processing. For the controller, we employ an exponential moving average of previous rewards with a weight of 0.95 as the baseline function in order to improve the performance and stability of RL. To encourage exploration, we add an entropy term into the loss with a weight of 0.01. We use the Adam optimizer for all MLP or RNN trainings. 
When mixing examples, we sample $\lambda \sim Beta(\alpha, \alpha)$ with $\alpha = 100$, which turns out to work well for regression tasks as we show in Sec.~\ref{sec:accuracyresults}. 
We generate the kNN options using linear or exponential series. For the possible numbers of NNs for {\sf NO2}, our default setting is the exponential series of integers $\{0\} \cup \{2^i | i\in [0, 7]\}$ (i.e., $|N| = 9$). For {\sf Bike}, {\sf Product}, and {\sf Synthetic}, the default series are {$\{0\} \cup \{4^i | i\in [0, 4]\}$}, $\{0\} \cup \{4^i | i\in [0, 4]\}$, and $\{0\} \cup \{2^i | i\in [0, 9]\}$, respectively. 

\paragraph{Baselines} 

We employ six baselines. First, we train a regression model on the labeled data without any data augmentation (``Vanilla''). Next, we faithfully implement the Mixup algorithm\,\citep{DBLP:conf/iclr/ZhangCDL18} where all example pairs can be mixed (``Mixup''). We also implement AdaMixup\,\cite{DBLP:conf/aaai/GuoMZ19} and Manifold Mixup\,\citep{verma2019manifold}, which are state-of-the-art Mixup techniques for classification that are also applicable to regression. We also compare with two simplified versions of \method{} where we use a global $k$ NN (``Global kNN'') or data distance $d$ threshold (``Global Distance'') for mixing examples with their nearest neighbors. The $k$ and $d$ parameters are tuned using Bayesian Optimization\,\citep{10.5555/646296.687872}.

\subsection{A Simple Evaluation on 1D Data}

\begin{figure}[t]
\centering
\includegraphics[width=0.5\columnwidth]{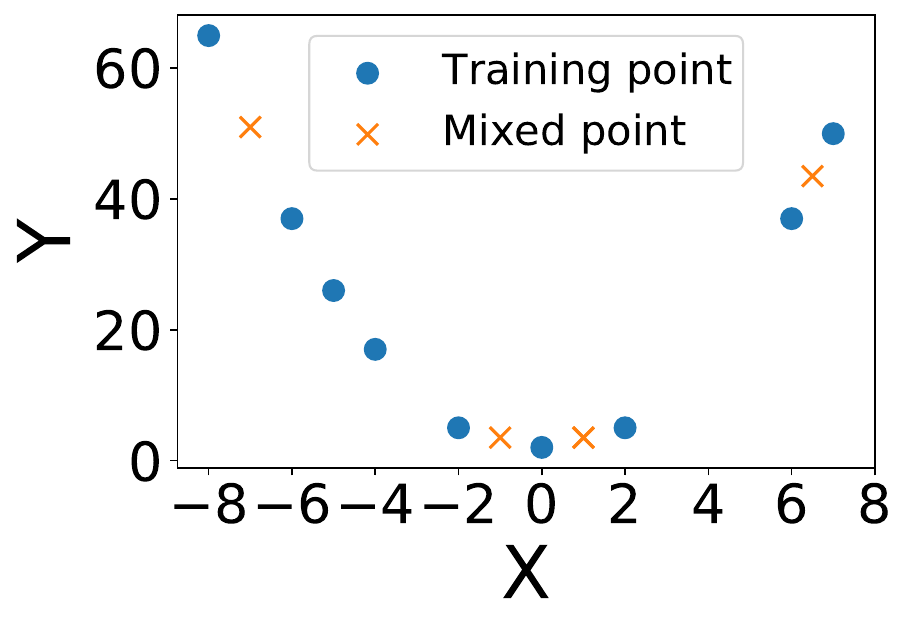}
\caption{\method{} results on 1D dataset. } 
\label{fig:1dmixing}
\end{figure}

As an exercise, we validate \method{}'s internal workings on an expanded 1-dimensional synthetic dataset based on Fig.~\ref{fig:comparison}. We generate 10 training examples using the same equation $y = \max\{x^2+1, 2\}$ with varying densities. We use the kNN options \{0, 1, 2, 4, 8\}. Fig.~\ref{fig:1dmixing} shows that \method{} mixes examples in ranges that are sparse, but not too sparse. The RMSE of a model trained on the data drops from 5.770 to 5.489 compared to not using \method{}.

\subsection{Data Distance Impact on Mixed Labels}
\label{sec:linearityexperiments}

\begin{figure}[t]
\begin{subfigure}[t]{0.45\columnwidth}
\centering
\includegraphics[scale=0.22]{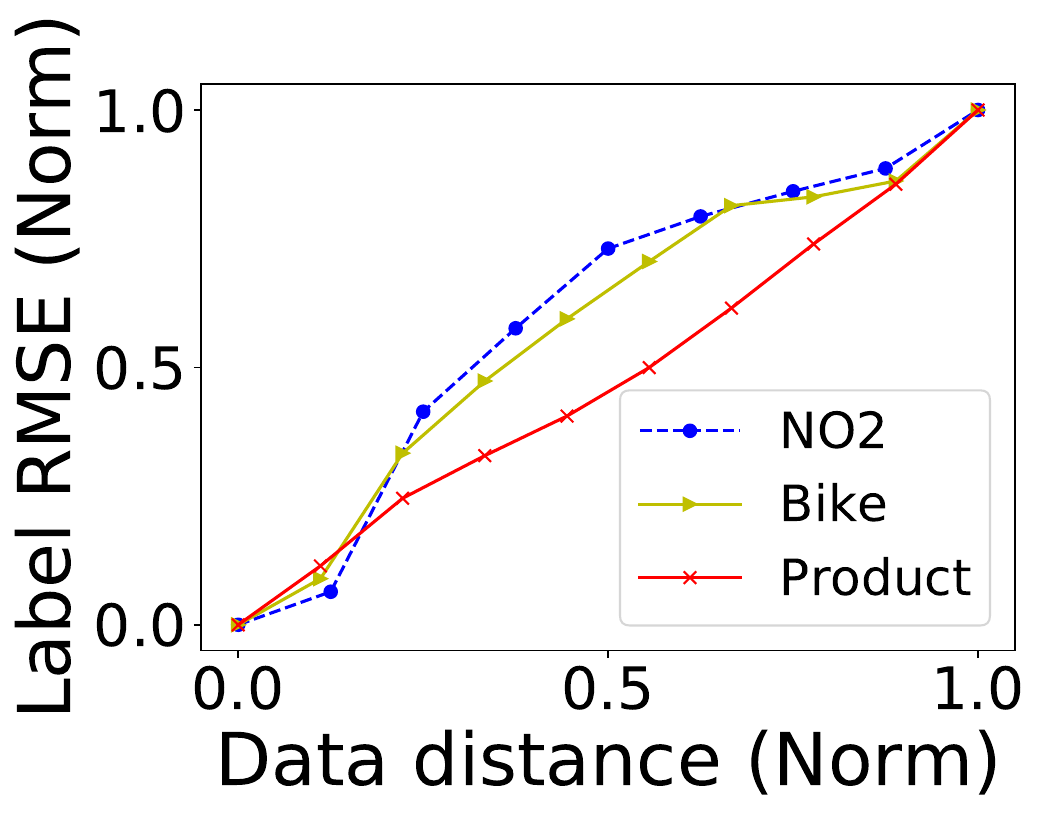}
\caption{Label RMSE}
\label{fig:labelerror1}
\end{subfigure}\hfill
\begin{subfigure}[t]{0.48\columnwidth}
\centering
\includegraphics[scale=0.22]{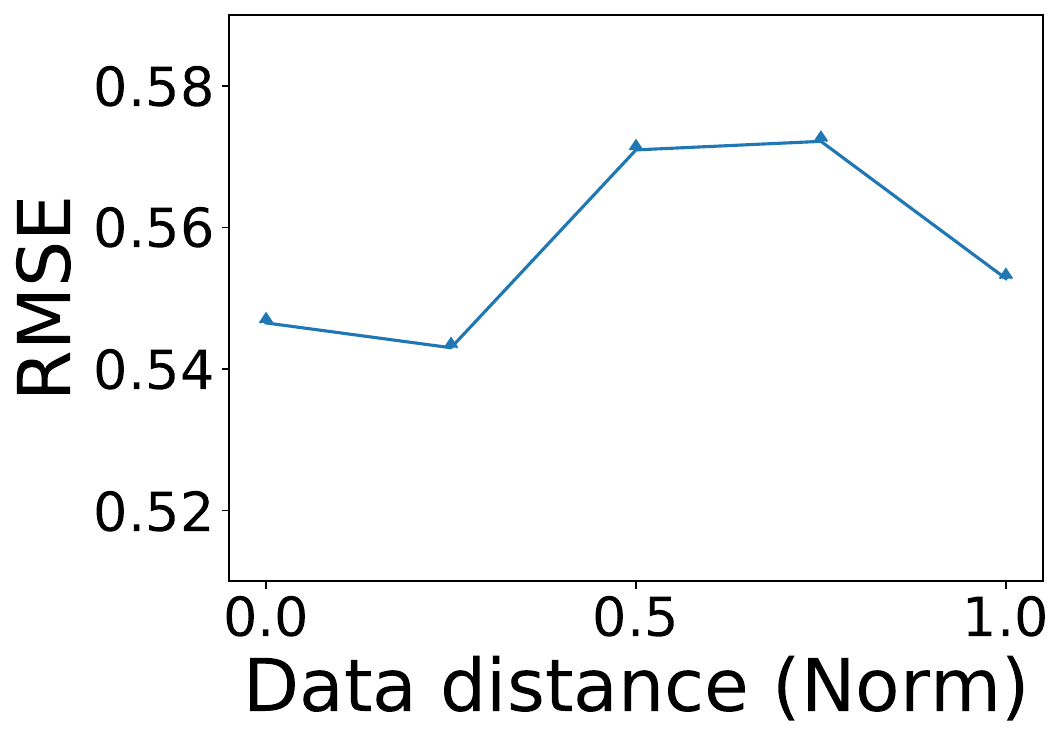}
\caption{{\sf NO2}}
\label{fig:datadistance_no2}
\end{subfigure}
\begin{subfigure}[t]{0.47\columnwidth}
\centering
\includegraphics[scale=0.22]{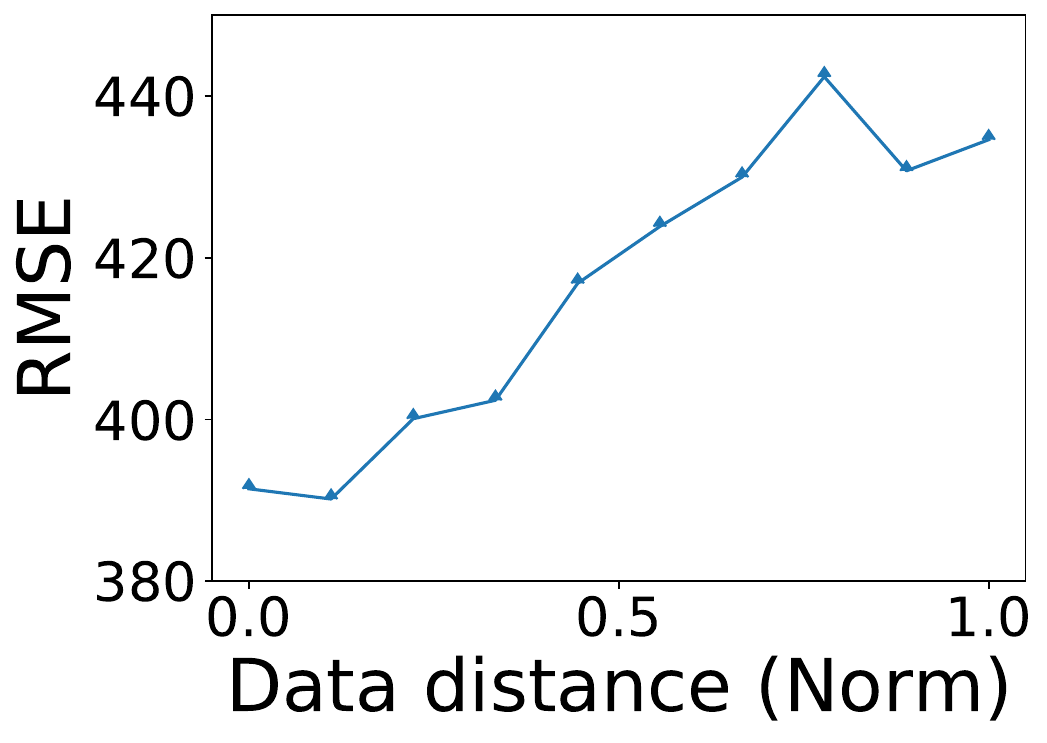}
\caption{{\sf Bike}}
\label{fig:datadistance_bike}
\end{subfigure} \hfill
\begin{subfigure}[t]{0.47\columnwidth}
\centering
\includegraphics[scale=0.22]{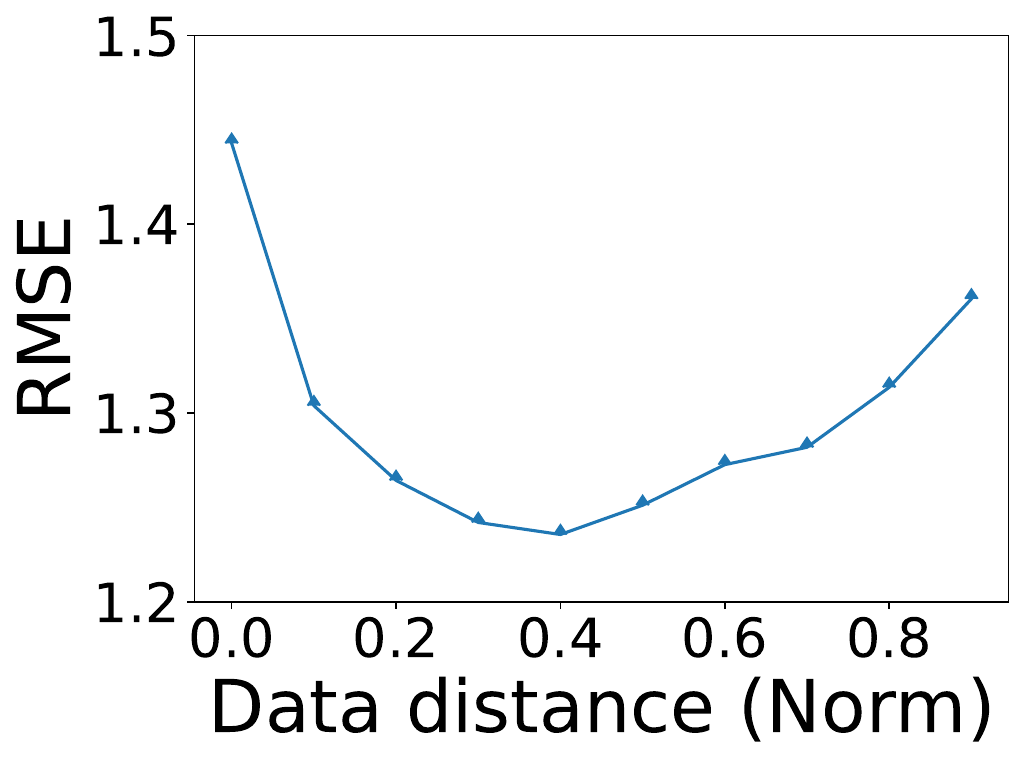}
\caption{{\sf Product}}
\label{fig:datadistance_product}
\end{subfigure}
  \caption{Label and Model RMSEs for the three real datasets (lower RMSE is better) when mixing examples with different ranges of (normalized) distances. (a) Mixing larger distances leads to higher label RMSEs. (b)-(d) Mixing larger distances also leads to higher model RMSE values.}
  \label{fig:datadistance}
\end{figure}

We empirically show that the greater the data distance between two examples, the more the interpolated labels are inconsistent with the distribution of the training data. Given a dataset, we first train a regression model. We then generate mixed examples and compute the RMSE between their interpolated labels and the labels predicted by the regression model. While the regression model is not perfect, the relative trend of the error gives sufficient insight. Fig.~\ref{fig:labelerror1} indeed shows increasing label errors for larger data distances for the three real datasets {\sf NO2}, {\sf Bike}, and {\sf Product}. We also train regression models on top of augmented training sets where we only mix examples within certain distance ranges. Figs.~\ref{fig:datadistance_no2}--\ref{fig:datadistance_product} show the model performances using RMSE for the three real datasets when mixing examples with different ranges of distances. Although the results vary by dataset, the overall trend is that increasing the data distance eventually causes the RMSE to increase to the extent that it is not worth mixing examples anymore.


\subsection{Performance Results}
\label{sec:accuracyresults}

\begin{table}[t]
  \centering
  \caption{RMSE results on all the real and synthetic datasets. Six baselines are compared with \method{}: (1) Vanilla; (2) Mixup\,\citep{DBLP:conf/iclr/ZhangCDL18}; (3) AdaMixup\,\citep{DBLP:conf/aaai/GuoMZ19}; (4) Manifold Mixup\,\citep{verma2019manifold}; (5) Global kNN; and (6) Global Distance.}
  \begin{tabular}{clc}
  \toprule
    Dataset & Method & RMSE \\
    \midrule
    \multirow{7}{*}{\sf NO2} & {Vanilla} & {$0.5441_{\pm0.0004}$} \\
    & {Mixup} & {$0.5401_{\pm0.0013}$} \\
    & {AdaMixup} & {$0.5397_{\pm0.0005}$} \\
    & {Manifold Mixup} & {$0.5381_{\pm0.0038}$} \\
    & {Global kNN (k=85)} & {$0.5470_{\pm0.0009}$} \\
    & {Global Dist. (d=0.81)} & {$0.5415_{\pm0.0017}$} \\
    & \method{} & {$\textbf{0.5248}_{\pm0.0015}$} \\
    \midrule
    \multirow{7}{*}{\sf Bike} & {Vanilla} & {$393.45_{\pm0.9049}$} \\
    & {Mixup} & {$393.36_{\pm1.8376}$} \\
    & {AdaMixup} & {$391.62_{\pm1.2527}$} \\
    & {Manifold Mixup} & {$399.44_{\pm0.7082}$} \\
    & {Global kNN (k=68)} & {$388.81_{\pm1.4053}$} \\
    & {Global Dist. (d=0.41)} & {$388.43_{\pm1.5282}$} \\
    & \method{} & {$\textbf{368.86}_{\pm3.0689}$} \\
    \midrule
    \multirow{7}{*}{\sf Product} & {Vanilla} & {$1.4100_{\pm0.0042}$} \\
    & {Mixup}  & {$1.2310_{\pm0.0107}$}  \\
    & {AdaMixup} & {$1.2293_{\pm0.0105}$} \\
    & {Manifold Mixup} & {$1.2894_{\pm0.0042}$} \\
    & {Global kNN (k=91)}  & {$1.2625_{\pm0.0102}$} \\
    & {Global Dist. (d=0.32)}  & {$1.2500_{\pm0.0172}$} \\
    & \method{} & {$\textbf{1.1948}_{\pm0.0023}$} \\
    \midrule
    \multirow{7}{*}{\sf Synthetic} & {Vanilla} & {$15.6838_{\pm0.2790}$} \\
    & {Mixup} & {$13.8358_{\pm0.0912}$} \\
    & {AdaMixup} & {$13.7602_{\pm0.0789}$} \\
    & {Manifold Mixup} & {$14.0457_{\pm0.1067}$} \\
    & {Global kNN (k=61)} & {$13.7587_{\pm0.0493}$} \\
    & {Global Dist. (d=0.35)} & {$14.0860_{\pm0.0875}$} \\
    & \method{} & {$\textbf{13.4935}_{\pm0.0913}$} \\
  \bottomrule
  \end{tabular}
  \label{tbl:regressionaccuracy}
\end{table}

\begin{figure}[t]
\begin{subfigure}[t]{0.22\textwidth}
\includegraphics[scale=0.22]{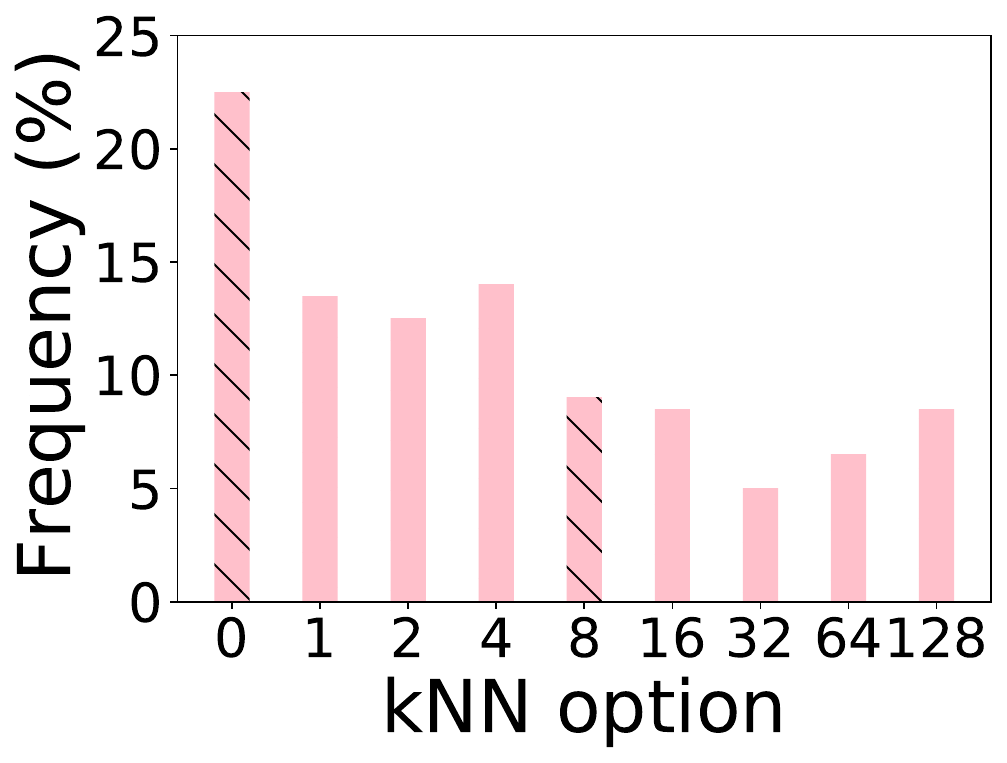}
\caption{{\sf NO2}}
\label{fig:knnhistogramno2}
\end{subfigure}\hfill
\begin{subfigure}[t]{0.22\textwidth}
\includegraphics[scale=0.22]{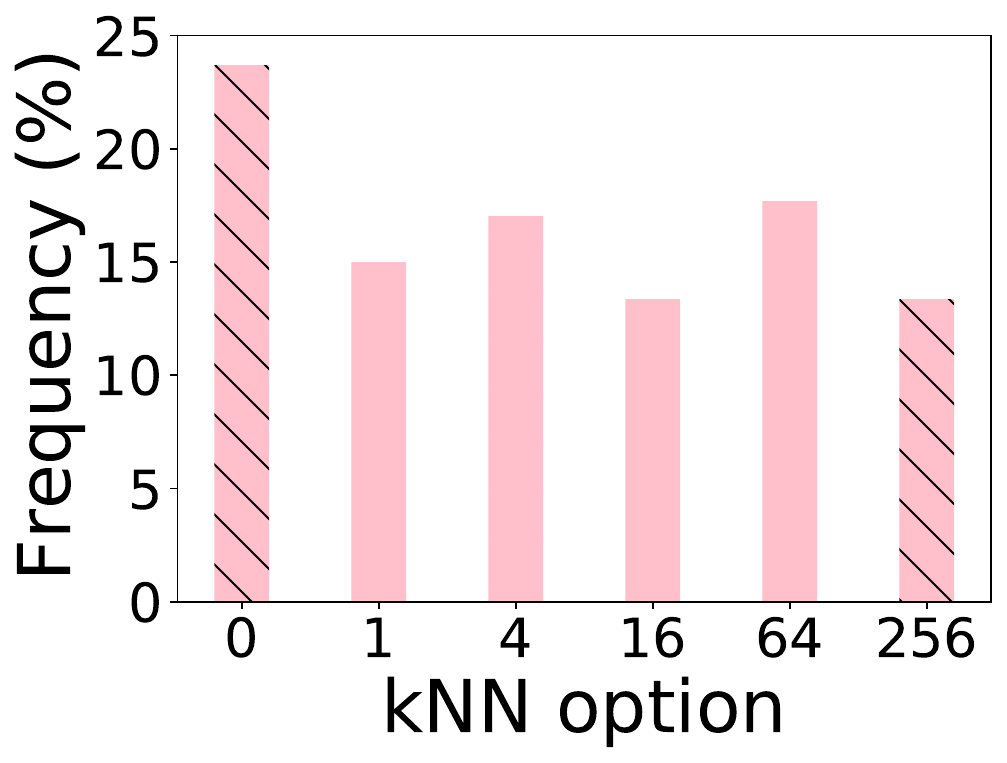}
\caption{{\sf Bike}}
\label{fig:knnhistogrambike}
\end{subfigure}
\begin{subfigure}[t]{0.22\textwidth}
\includegraphics[scale=0.22]{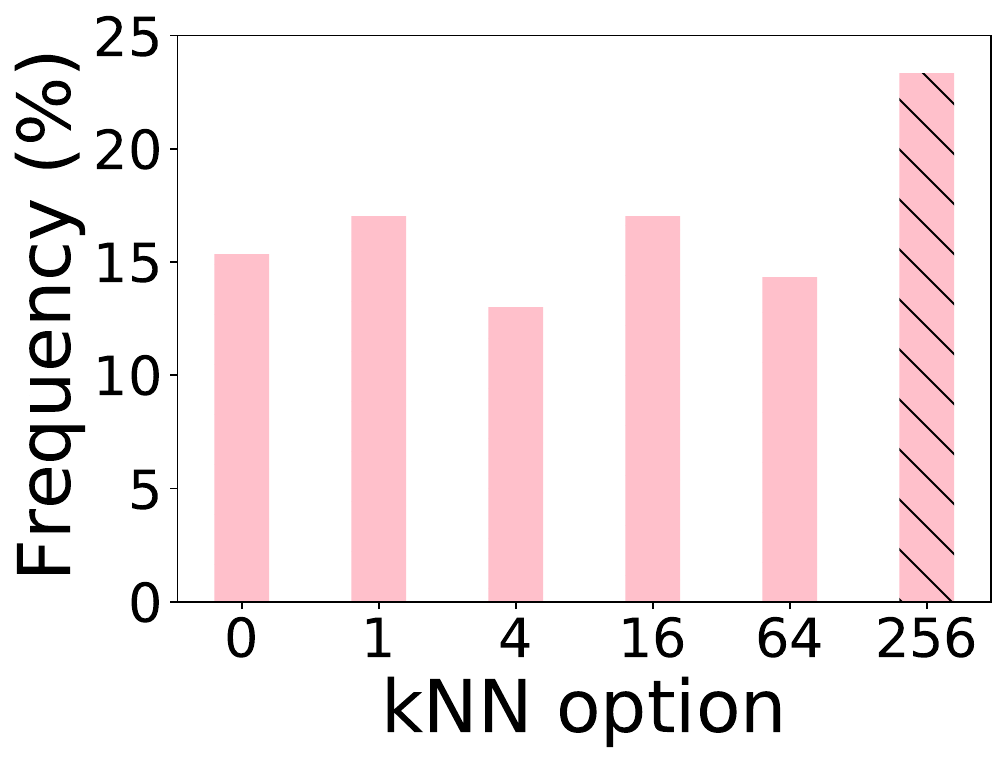}
\caption{{\sf Product}}
\label{fig:knnhistogramproduct}
\end{subfigure} \hfill
\begin{subfigure}[t]{0.22\textwidth}
\includegraphics[scale=0.22]{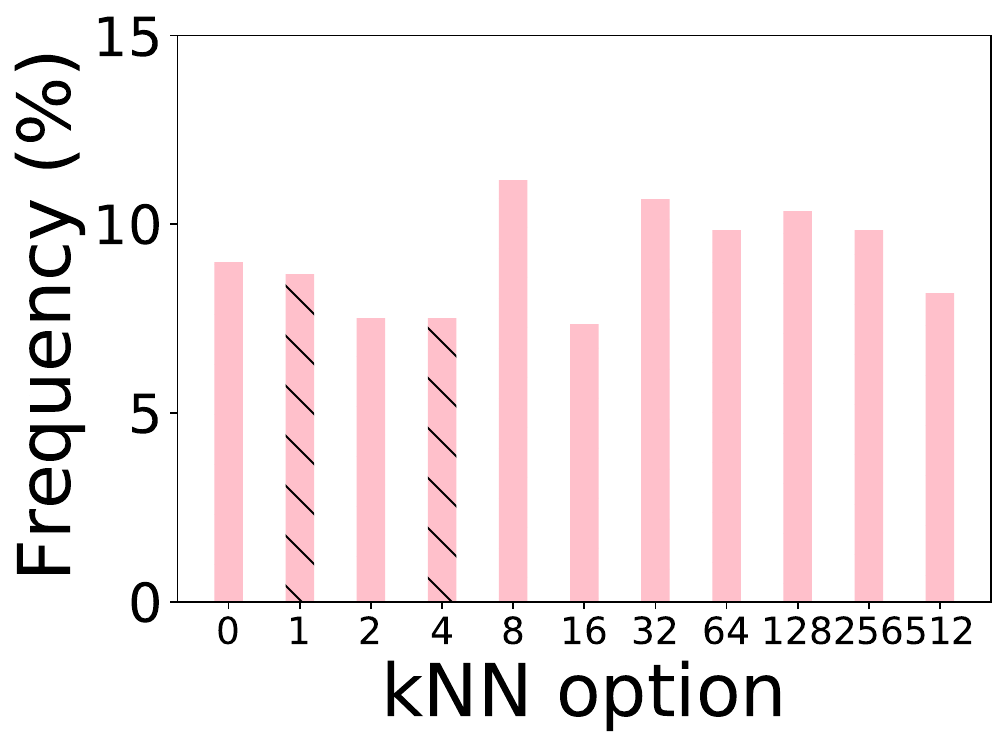}
\caption{{\sf Synthetic}} 
\label{fig:knnhistogramsynthetic}
\end{subfigure} 
  \caption{kNN option frequency histograms for all datasets.} 
    \label{fig:knnhistograms}
\end{figure}

We compare \method{} with the six baselines using the settings in Table~\ref{tbl:datasetcomparison}. Table~\ref{tbl:regressionaccuracy} shows the results for all the datasets. \method{} consistently outperforms all the baselines in terms of regression model performance, which demonstrates that it is able to pinpoint which examples should be mixed with how many kNNs. 
Fig.~\ref{fig:knnhistograms} shows which kNN options are frequently used by \method{} on all the datasets. When reading the histograms, keep in mind that the x-axis is an exponential series instead of a linear one. The y-axis is the frequency ratio ranging from 0 to 1. For Fig.~\ref{fig:knnhistogramno2}, the frequency ratios are high for small kNN values and gradually lower for larger kNN values. This result is consistent with Fig.~\ref{fig:datadistance_no2} where mixed examples with the longer distance increase RMSE. Fig.~\ref{fig:knnhistogrambike} has a similar distribution as Fig.~\ref{fig:knnhistogramno2} and is consistent with Fig.~\ref{fig:datadistance_bike}. Fig.~\ref{fig:knnhistogramproduct}'s histogram is more uniform, which means that the larger kNN values have relatively higher frequencies compared to the {\sf NO2} and {\sf Bike} datasets. This result is consistent with Fig.~\ref{fig:datadistance_product}, where mixing is actually helpful for a wide range of data distances. Finally, Fig.~\ref{fig:knnhistogramsynthetic} has a similar distribution as Fig.~\ref{fig:knnhistogramproduct}.

\begin{figure}[t]
\begin{subfigure}[t]{0.23\textwidth}
\centering
\includegraphics[scale=0.23]{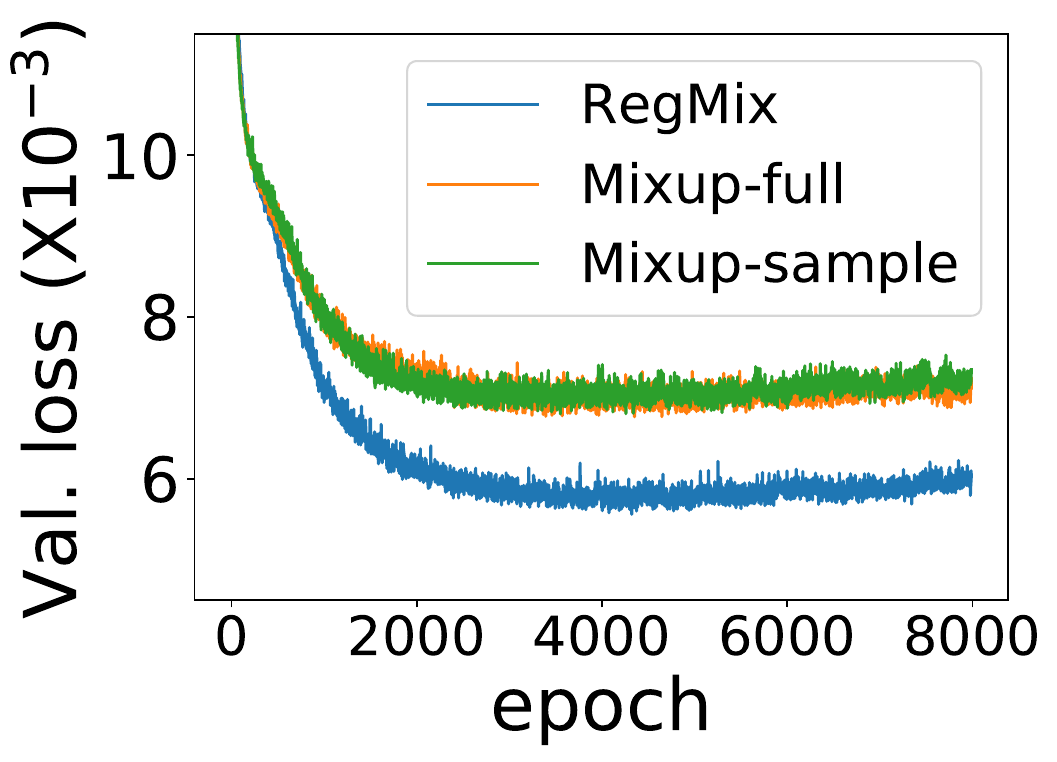}
\caption{Training convergence}
\label{fig:convergencecomparison}
\end{subfigure}
\begin{subfigure}[t]{0.23\textwidth}
\centering
\includegraphics[scale=0.23]{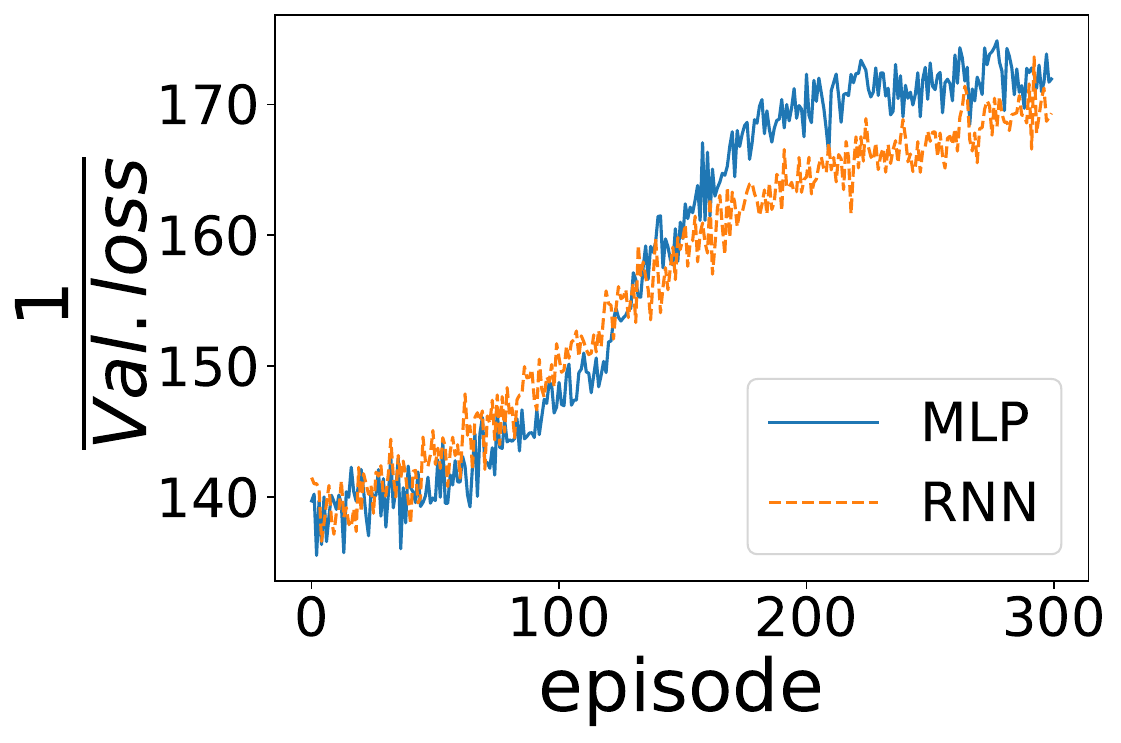}
\caption{Reward}
\label{fig:validationaccuracy}
\end{subfigure}

  \caption{(a) Model training convergence comparison on the {\sf Product} dataset. (b) Rewards when using the MLP and RNN controllers on the {\sf Product} dataset.} 
    \label{fig:}
\end{figure}

We further compare \method{} and Mixup on the {\sf Product} dataset by plotting the model training convergence speeds on top of the augmented data (Fig.~\ref{fig:convergencecomparison}). For Mixup, we evaluate two scenarios: when we use the entire mixed data or take a random sample of the mixed data that has the same size as \method{}'s augmented data for a fair comparison. As a result, the convergence is much faster when the model is trained on \method{}'s augmented data.

\paragraph{K Nearest Neighbor Options}

We investigate how changing the kNN options $N$ affects \method{}'s performance in Table~\ref{tbl:knnoptions}. For each dataset, we compare its default series with other possible series. We also fix the entire runtime by using the same number of epochs, so we only need to compare the model performances. We observe that providing more kNN options improves \method{}'s performance, but only to a certain extent. Adding options unnecessarily may actually result in lower performances as the best options are more difficult to find within the same time. Hence, \method{} performs best with a modest number of options.

\begin{table}[t]
  \centering
  \caption{\method{} performances when using different kNN options on all datasets.}
  \begin{tabular}{@{\hspace{4pt}}c@{\hspace{4pt}}l@{\hspace{4pt}}c@{\hspace{4pt}}c@{\hspace{4pt}}} 
  \toprule
    Dataset & kNN Options (Default = $\dagger$) & RMSE \\
    \midrule
    \multirow{3}{*}{\sf NO2} & {$\{0\} \cup \{4^i | i\in [0, 3]\}$} & {$0.5357_{\pm0.0020}$} \\
    & {$\{0\} \cup \{2^i | i\in [0, 7]\}^\dagger$} & {$0.5248_{\pm0.0015}$} \\
    & {$\{0\} \cup \{10i | i\in [1, 19]\}$} & {${0.5317}_{\pm0.0011}$} \\
    \midrule
    \multirow{3}{*}{\sf Bike} & {$\{0\} \cup \{8^i | i\in [0, 2]\}$} & {$378.02_{\pm2.6582}$} \\
    & {$\{0\} \cup \{4^i | i\in [0, 4]\}^\dagger$} & {$368.86_{\pm3.0689}$} \\
    & {$\{0\} \cup \{2^i | i\in [0, 8]\}$} & {$370.68_{\pm0.6302}$} \\
       
    \midrule
    \multirow{3}{*}{\sf Product} & {$\{0\} \cup \{8^i | i\in [0, 2]\}$} & {$1.2276_{\pm0.0051}$} \\
    & {$\{0\} \cup \{4^i | i\in [0, 4]\}^\dagger$} & {$1.1948_{\pm0.0023}$} \\
    & {$\{0\} \cup \{2^i | i\in [0, 8]\}$} & {$1.1954_{\pm0.0100}$} \\

    \midrule
    \multirow{3}{*}{\sf Synthetic} & {$\{0\} \cup \{8^i | i\in [0, 3]\}$} & {$13.7498_{\pm0.0449}$} \\
    & {$\{0\} \cup \{2^i | i\in [0, 9]\}^\dagger$}  & {${13.4935}_{\pm0.0913}$} \\
    & {$\{0\} \cup \{30i | i\in [1, 19]\}$} & {$13.5510_{\pm0.0916}$} \\
  \bottomrule
  \end{tabular}
  \label{tbl:knnoptions}
\end{table}

\paragraph{Comparison with RNN Controller}

We compare \method{} using the default MLP controller with \method{} that uses the RNN controller in Table~\ref{tbl:rnncontroller} using the three real datasets. To clearly observe the RNN controller's performance depending on the example ordering in the RNN, we consider two scenarios where (1) the examples are ordered after dimensionality reduction to one dimension using PCA\,\citep{doi:10.1080/14786440109462720} and (2) the examples are ordered randomly. As a result, there is no significant difference in model performances among the three approaches, which suggests that any influence among examples when mixing with neighbors is either minor or difficult to capture. As discussed in Sec.~\ref{sec:controller}, we suspect that there is no easy way to model the influence by ordering the examples. Fig.~\ref{fig:validationaccuracy} shows that the reward trends for the two controllers are similar as well. 


\begin{table}[t]
  \centering
  \caption{\method{} performances when using an MLP or RNN controller on the three real datasets. When using an RNN controller, we either (1) sort the examples using dimensionality reduction or (2) order them randomly.}
  \begin{tabular}{clc}
  \toprule
    Dataset & Controller & RMSE \\
    \midrule
    \multirow{3}{*}{\sf NO2} & MLP & {$0.5248_{\pm0.0015}$} \\
    & RNN (dim. reduction) & {$0.5277_{\pm0.0031}$} \\
    & RNN (rand. selection) & {$0.5279_{\pm0.0004}$} \\
    \midrule
    \multirow{3}{*}{\sf Bike} & MLP & {$368.86_{\pm3.0689}$}  \\
    & RNN (dim. reduction) & {$368.03_{\pm3.5014}$} \\
    & RNN (rand. selection) & {$370.22_{\pm1.3731}$} \\
    \midrule
    \multirow{3}{*}{\sf Product} & MLP & {$1.1948_{\pm0.0023}$} \\
    & RNN (dim. reduction) & {$1.1943_{\pm0.0065}$} \\
    & RNN (rand. selection) & {$1.1963_{\pm0.0075}$} \\
  \bottomrule
  \end{tabular}
  \label{tbl:rnncontroller}
\end{table}

\paragraph{Varying $\alpha$}

\begin{figure}[t]
\begin{subfigure}[t]{0.23\textwidth}
\centering
\includegraphics[scale=0.23]{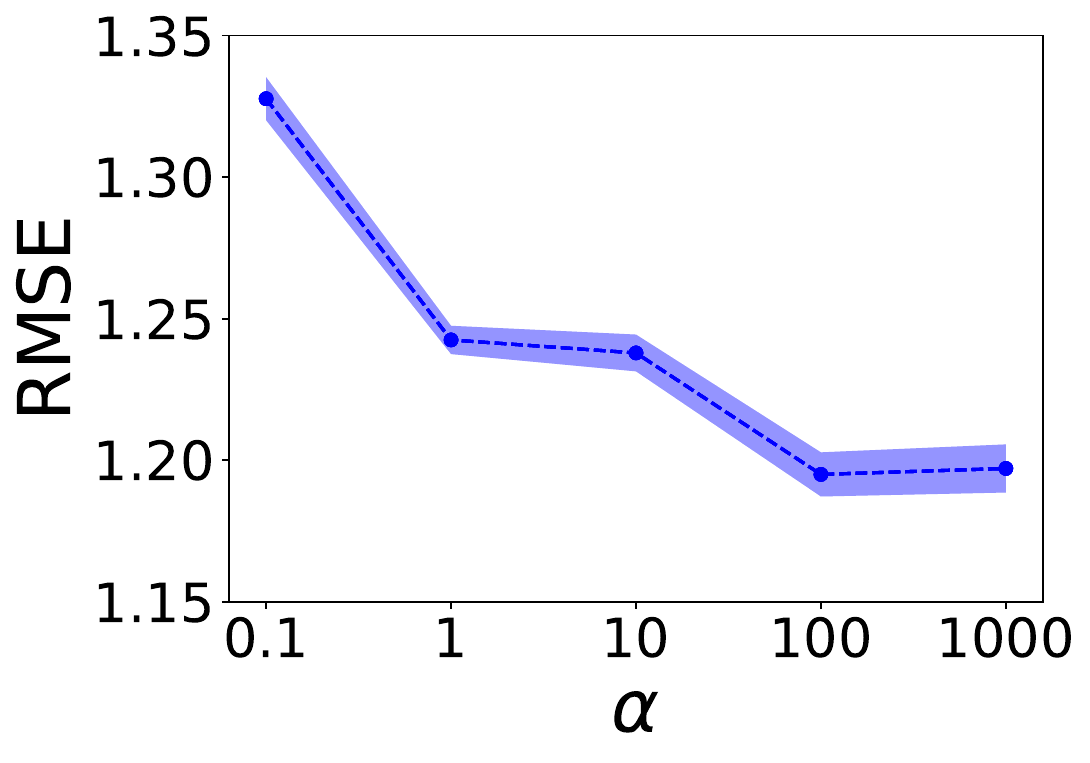}
\caption{$\alpha$ impact}
\label{fig:lambda_effect_product}
\end{subfigure}
\begin{subfigure}[t]{0.23\textwidth}
\centering
\includegraphics[scale=0.23]{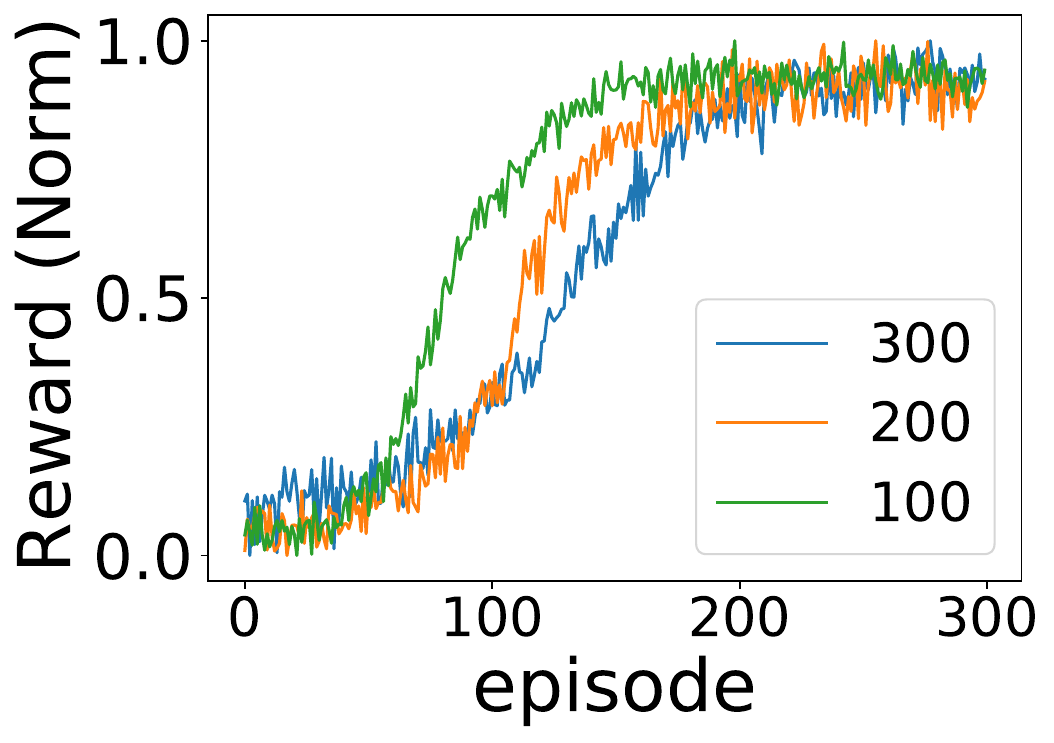}
\caption{Convergence speed}
\label{fig:convergence_size}
\end{subfigure}
\caption{(a) \method{} performances using different $\alpha$ values on the {\sf Product} dataset. (b) \method{}'s reward convergences on different amounts of {\sf Product} data.} 
    \label{fig:}
\end{figure}

In Fig.~\ref{fig:lambda_effect_product}, we vary $\alpha$ to adjust the Beta distribution from which we sample $\lambda$ for the {\sf Product} dataset. As a result, an $\alpha$ of at least 100 results in the best model performances where $\lambda$ is very close to 0.5. The results for the other real datasets are similar and shown in the supplementary. We suspect that generating examples in the middle of two examples helps the regularization the most for regression. 


\paragraph{Other Regression Models}

We also evaluate \method{} using a LightGBM\,\cite{DBLP:conf/nips/KeMFWCMYL17} model (Table~\ref{tbl:regnomlp}) and observe similar performance improvements as in Table~\ref{tbl:regressionaccuracy}. Hence, \method{} is also effective on models beyond MLPs.

\begin{table}[t]
  \centering
  \caption{\method{} using a LightGBM\,\cite{DBLP:conf/nips/KeMFWCMYL17} model.}
  \begin{tabular}{cclc}
  \toprule
    Dataset & Model & Method & RMSE \\
    \midrule
    \multirow{3}{*}{\sf NO2} & \multirow{3}{*}{LightGBM} & Vanilla & {$0.5334$} \\
    & & Mixup & {$0.5467$} \\
    & & \method{} & {$\textbf{0.5258}$} \\
    \midrule
    \multirow{3}{*}{\sf Bike} & \multirow{3}{*}{LightGBM} & Vanilla & {$385.31$} \\
    & & Mixup & {$335.15$} \\
    & & \method{} & {$\textbf{311.84}$} \\
  \bottomrule
  \end{tabular}
  \label{tbl:regnomlp}
\end{table}

\subsection{Convergence Speed and Post-hoc Method}

We evaluate the convergence speed of \method{} on different amounts of {\sf Product} data (Fig.~\ref{fig:convergence_size}). The number of epochs needed for convergence increases for larger dataset sizes. In terms of runtime, \method{} takes hours to converge. Although \method{} can be viewed as expensive, the gained model performance can be worth the effort in real applications because data augmentation is a one-time cost that can run in batch mode. We discuss how to possibly improve \method{}'s runtime in the supplementary.


\method{} can also be used as a ``post-hoc'' step in model development where one first trains a model including hyperparameter tuning and then further train the model on the augmented data from \method{} using the same hyperparameters, if an extra gain in model performance is critical. We perform the same experiments as in Table~\ref{tbl:regressionaccuracy} running \method{} post-hoc and observe that the RMSE values for \method{} are near identical as those in Table~\ref{tbl:regressionaccuracy}: 0.5286 for {\sf NO2}, 375.57 for {\sf Bike}, 1.2073 for {\sf Product}, and 13.7943 for {\sf Synthetic}.

\section{Related Work}
\label{sec:relatedwork}



Data augmentation is essential for model performance\,\cite{DBLP:conf/interspeech/ParkCZCZCL19,DBLP:conf/iclr/ZhangCDL18,DBLP:conf/nips/KrizhevskySH12,DBLP:conf/icdar/SimardSP03}, but there has been much more emphasis on classification than regression. There are largely three approaches: generative models, policies, and Mixup techniques. Generative models including GANs\,\citep{DBLP:conf/nips/GoodfellowPMXWOCB14} and VAEs\,\citep{DBLP:journals/corr/KingmaW13} are popular in classification where the idea is to generate realistic data that cannot be distinguished from the real data by a discriminator. However, a major assumption is that the labels of the generated examples are the same, which does not necessarily hold in a regression setting where most examples may have different labels. Another approach is to find policies\,\citep{DBLP:conf/cvpr/CubukZMVL19,DBLP:conf/nips/RatnerEHDR17,DBLP:conf/icml/HoLCSA19,DBLP:conf/cvpr/CubukZSL20,DBLP:conf/nips/LimKKKK19,hataya2020a}, which specify fixing rules for transforming the data while maintaining the label value. For example, image processing policies may flip, rotate, or adjust the brightness of images to generate new valid images\,\cite{DBLP:conf/bmvc/ZagoruykoK16,DBLP:conf/nips/KrizhevskySH12}. However, these policies usually target image classification or object detection tasks. In a regression setting, the same transformed examples may now have different unknown labels, making them unsuitable for training. 

Mixup\,\citep{verma2019manifold, DBLP:conf/iccv/YunHCOYC19, kim2020puzzle, kim2021comixup} takes the alternative approach of generating both data and labels together by mixing existing examples with different labels using linear interpolations\,\citep{DBLP:conf/nips/ChapelleWBV00,DBLP:conf/icml/WuZVR20}. Mixup can also be used in a regression setting, and there are analytical results on how it is effective by regularizing the model\,\citep{DBLP:journals/corr/abs-2006-06049, DBLP:conf/iclr/ZhangDKG021}. However, our key observation is that mixing distant examples may actually be detrimental to model performance. \method{} shows better performance than Mixup by carefully choosing for each example which nearest neighbors to mix.

Within reinforcement learning\,\citep{DBLP:books/lib/SuttonB18,10.5555/1622737.1622748}, we are interested in policy searching methods\,\citep{neumann2015policysearch} where our goal is to find the best policy for mixing examples. In particular, \method{}'s framework is inspired by the recent AutoAugment framework\,\citep{DBLP:conf/cvpr/CubukZMVL19,DBLP:conf/nips/LimKKKK19, hataya2020a}, which uses PPO\,\citep{DBLP:journals/corr/SchulmanWDRK17} to search for data augmentation policies that dictate how to modify existing examples to generate additional training data for image classification. In comparison, we solve the completely different problem of data augmentation for regression where we extend Mixup and find the best kNN neighbors to mix for each example.

\section{Conclusion}
\label{sec:conclusion}

We proposed \method{}, which is to our knowledge the first Mixup-based data augmentation framework for regression tasks utilizing distance information. We observe that mixing distant examples may not be beneficial and even detrimental to model performance. Hence, \method{} extends Mixup by determining the nearest neighbors to mix for each example using reinforcement learning where the objective is to minimize the regression model's loss on a validation set. In our experiments, we showed that \method{} outperforms various data augmentation baselines for regression by effectively selecting among a modest number of nearest neighbor options. 



\section{Acknowledgment}
Sung Yoon Ryu, Soo Seok Lee, and Gwang Nae Gil of Samsung Electronics provided the Product dataset and key insights on the 3-d semiconductor application. This work is supported by the Samsung Electronics’ University R$\&$D program [Algorithm advancement and reliability-enhancing technique development for spectrum data-based deep learning] and a Kwon Oh-Hyun Assistant Professorship.

\bibliography{main}



\clearpage
\appendix
\section{Appendix}

\subsection{Mixup Analysis for Cont. Regression Model}

Suppose that the regression model $f$ is continuous where $\lim_{x \rightarrow c} f(x) = f(c)$ for any $x$ and $c$ within the domain of $f$. We show that a short-enough data distance sufficiently reduces the label distance as well. Given $f$'s domain $D$ and $\lim_{x \rightarrow c} f(x) = L$, the following is known to hold: $\forall \epsilon, \exists \delta$ s.t. $\forall x \in D$, if $|x - c| < \delta$, then $|f(x) - L| < \epsilon$. We can use this result to prove that $\forall \epsilon, \exists \delta$ s.t. $\forall x_i, x_j \in D$, if $\alpha x_i + (1 - \alpha) x_j = c$, $0 \leq \alpha \leq 1$, and $|x_i - x_j| < \delta$ then the absolute difference between the mixed example's $x_j$ value and $L$ is small where $|\alpha f(x_i) + (1 - \alpha) f(x_j) - L| = |\alpha (f(x_i) - L) + (1 - \alpha) (f(x_j) - L)| \leq \alpha |f(x_i) - L| + (1 - \alpha) |f(x_j) - L| < \alpha \epsilon + (1 - \alpha) \epsilon = \epsilon$. While this analysis does not cover all edge cases (e.g., identical examples having different labels), it does give an intuition why limiting the data distances for mixing is a reasonable solution.

\subsection{More Experimental Settings}
\label{appendix:moreexperimentalsettings}

We continue describing our experimental settings from Sec.~\ref{sec:experiments}. 
When constructing the {\sf Synthetic} dataset, recall that we take a subset of the entire DACON challenge dataset\,\citep{dacon}. While the the range of thicknesses for the full datasets is [10, 300] for all four layers, we only use examples with thicknesses within the range [10, 100]. The purpose is to reduce the training time while still making a clear comparison between \method{} and the other methods.




\subsection{More Experiments on $\alpha$ Impact}
\label{appendix:alphaimpact}
\begin{figure}[h]
\begin{subfigure}[t]{0.23\textwidth}
\centering
\includegraphics[scale=0.23]{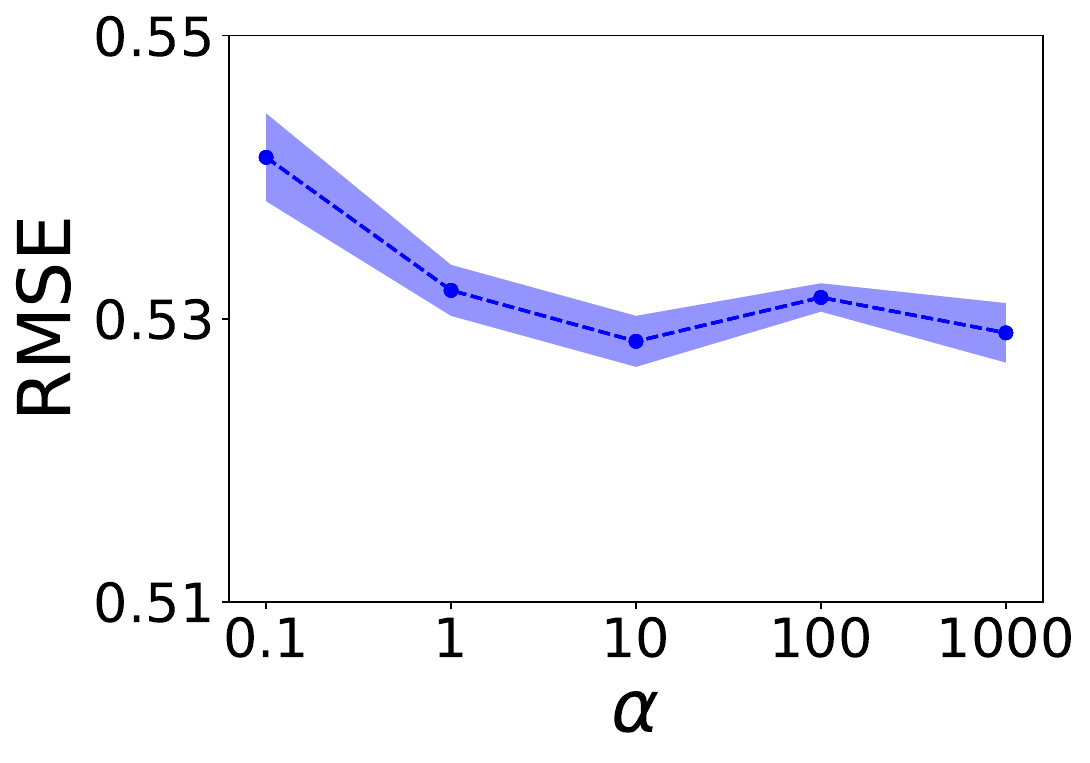}
\caption{$\sf NO2$}
\label{fig:alpha_effect_no2}
\end{subfigure}
\begin{subfigure}[t]{0.23\textwidth}
\centering
\includegraphics[scale=0.23]{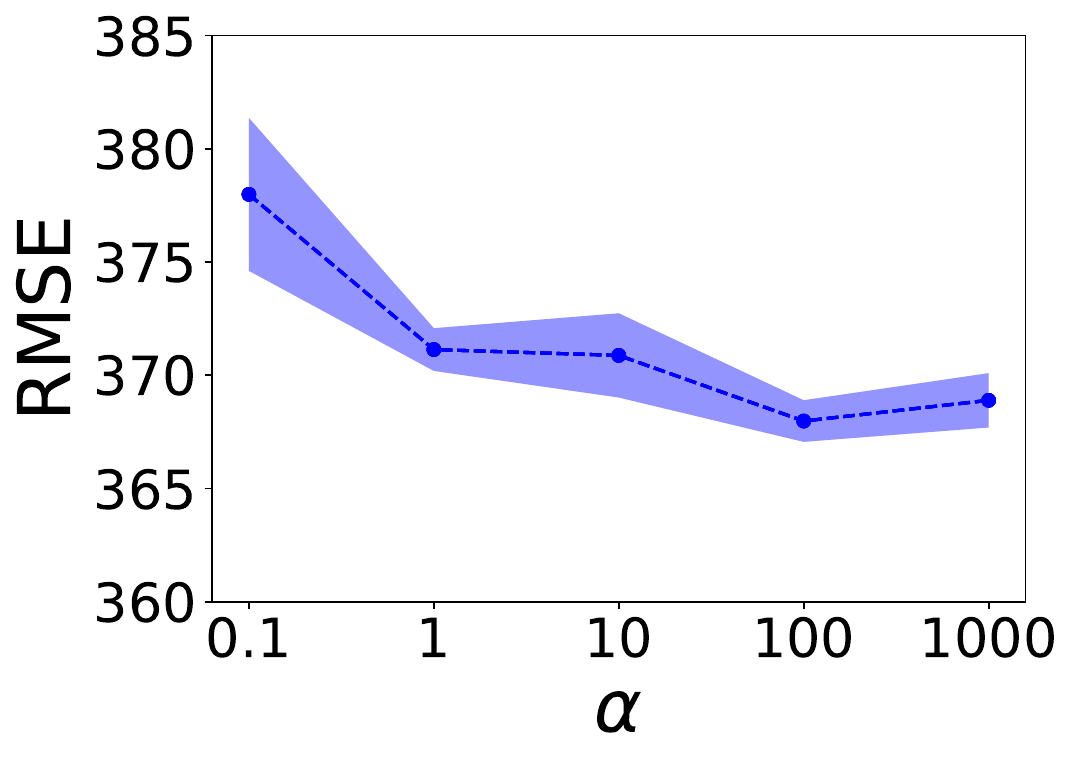}
\caption{$\sf Bike$}
\label{fig:alpha_effect_bike}
\end{subfigure}
\caption{\method{} performances for different $\alpha$ values of the Beta distribution for sampling $\lambda$ on (a) {\sf NO2} and (b) {\sf Bike} datasets.} 
    \label{fig:}
\end{figure}

We continue showing \method{} performances for different $\alpha$ values of Beta distribution on {\sf NO2} and {\sf Bike} datasets in Fig~\ref{fig:alpha_effect_no2} and Fig~\ref{fig:alpha_effect_bike}. As a result, using an $\alpha$ of at least 100 results in the lowest RMSE.

\subsection{Improving \method{}'s Runtime}

We would also like to further improve \method{}'s runtime by possibly adapting techniques from efficient versions of AutoAugment like Fast AutoAugment\,\cite{DBLP:conf/nips/LimKKKK19} or Faster AutoAugment\,\cite{hataya2020a}. These methods view data augmentation as filling missing data points in the training data and skip the iterative full training process. However, the methods require large amounts of training data to train a model that accurately estimates the training data distribution. In our setup where the training data is small, the trained distribution will overfit to the training data, making these techniques less effective. Utilizing other scalable techniques is an interesting future work.

\end{document}